\documentclass[runningheads]{llncs}
\usepackage{graphicx}
\usepackage{amsmath,amssymb} 
\usepackage{color}
\usepackage[width=122mm,left=12mm,paperwidth=146mm,height=193mm,top=12mm,paperheight=217mm]{geometry}
\begin{document}
\pagestyle{headings}
\mainmatter
\def\ECCV16SubNumber{5752}  

\title{BTranspose: Bottleneck Transformers for Human Pose Estimation with Self-Supervised Pre-Training} 

\titlerunning{BTranspose: Human Pose Estimation}

\authorrunning{ECCV-22 submission ID \ECCV16SubNumber}

\author{Kaushik Balakrishnan\inst{1} \and Devesh Upadhyay\inst{2}}

\institute{Ford Greenfield Labs, Palo Alto, CA, \email{kbalak18@ford.com} \and Ford Research, Dearborn, MI, \email{dupadhya@ford.com}}

\maketitle

\begin{abstract}
   The task of 2D human pose estimation is challenging as the number of keypoints is
typically large ($\sim$ 17) and this necessitates the use of robust 
neural network architectures and training pipelines that can capture 
the relevant features from the input image. These features are then aggregated to 
make accurate heatmap predictions from which the final keypoints of
human body parts can be inferred. Many papers in literature use 
CNN-based architectures for the backbone, and/or combine it with a 
transformer, after which the features are aggregated to make the final
keypoint predictions \cite{Transpose}. In this paper, we consider the recently proposed 
Bottleneck Transformers \cite{BottleneckTransformers}, which combine CNN and multi-head self attention (MHSA) layers
effectively, and we integrate it with a Transformer encoder and apply it to the task of 2D human pose estimation.
We consider different backbone architectures and pre-train them using 
the DINO self-supervised learning method \cite{DINO}, this pre-training is found to improve the 
overall prediction accuracy. 
We call our model BTranspose, and experiments show that on the {\it COCO}
validation set, our model achieves an AP of 76.4, which is competitive with 
other methods such as \cite{Transpose} and has fewer network parameters.
Furthermore, we also present the dependencies of the final predicted keypoints on 
both the MHSA block and the Transformer encoder layers, providing clues on the image sub-regions the
network attends to at the mid and high levels. 
\end{abstract}

\section{Introduction}

\begin{figure*}[!h]
\centering%
(a)\includegraphics[width=5cm]{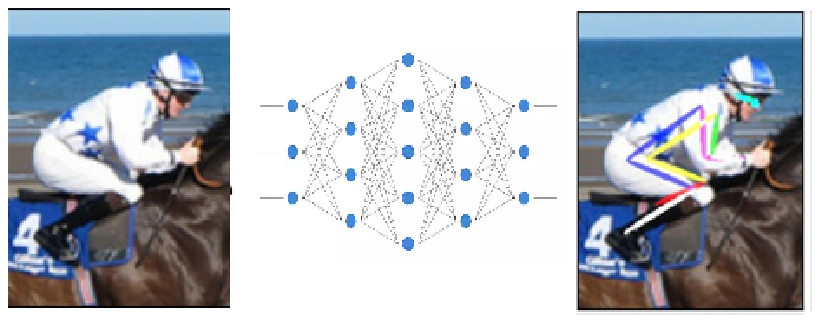} (b)\includegraphics[width=5cm]{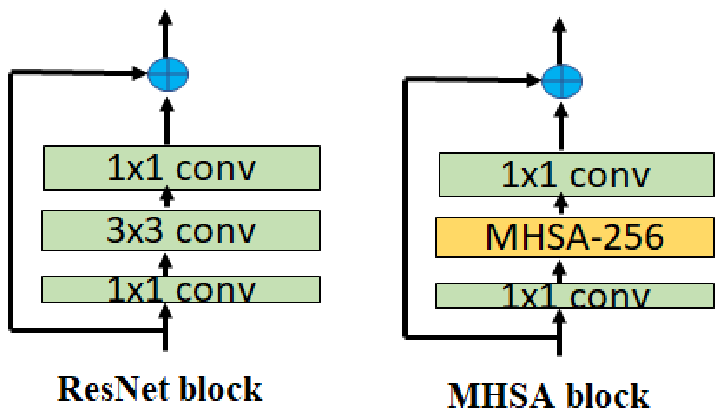} \\
\caption{(a) Schematic of the pose estimation problem. (b) Bottleneck Transformers: comparison of the ResNet block \cite{Resnet} and the MHSA block \cite{BottleneckTransformers}.}
\label{fig:bottleneck}%
\end{figure*}

Deep convolutional neural network (CNN) architectures have a proven track
record of success in image classification \cite{Lenet}, \cite{Alexnet}, object detection \cite{Girshick2014}, \cite{Fastrcnn}, \cite{COCO}, semantic segmentation \cite{Deeplabv3}, \cite{Gscnn}, etc. 
CNNs are also popular in human pose estimation. 
For instance, DeepPose \cite{Deeppose} regresses the numerical coordinate locations of keypoints. 
Fully convolutional networks such as \cite{Newell2016}, \cite{Chen2018}, \cite{Papandreou2017}, \cite{personlab}, \cite{Xiao2018}
predict keypoint heatmaps, which implicitly learn the dependencies between the human body parts.
All these studies rely on a CNN backbone, which unfortunately makes it
unclear how to explain the network's learning of the relationships between the different human body parts.
Moreover, the CNN models used for human pose estimation \cite{Newell2016}, \cite{Chen2018}, \cite{Papandreou2017}, \cite{personlab}, \cite{Xiao2018}
have very deep architectures, which makes it cumbersome to ascertain what each layer of the network learns and how it is related to the 
final output predictions. 
A schematic of the human keypoint detection is shown in Fig. \ref{fig:bottleneck} (a). 
The task of keypoint detection inherently involves learning long-range dependencies: the network must 
collect and associate body parts of the human in an image, and for this to be accurately learned, the 
network must learn the relationships across sub-regions of the image, which is challenging to undertake in
purely convolutional neural network architectures. 
Convolutional-based architectures require stacking multiple layers \cite{Simonyan2014}, \cite{Resnet}
in order to efficiently learn the local-to-global relationships.   

Learning the long-range dependencies is critical for
robust predictions of human pose. In natural language processing (NLP) applications, 
self-attention has a proven track record of success as the basic building block of 
Transformer architectures to learn such local-to-global correspondences \cite{Vaswani2017}, \cite{BERT}, including for 
long sequences.
The self-attention building block has also been extended to computer vision 
applications by stacking Transformer blocks---the Vision Transformer (ViT)---by feeding image patches into the Transformer \cite{Dosovitskiy2020}
and learning the relationships between the patches. This demonstrated success of 
self-attention as  the fundamental building block (i.e., in the Transformer)
has made it an alternative to the more classical convolutional architectures for many applications.
Recently, the Bottleneck Transformer \cite{BottleneckTransformers} was introduced 
by blending CNN and MHSA layers. The standard ResNet \cite{Resnet} consists of multiple ``ResNet blocks"
where each of the block is a sequence of 1$\times$1, 3$\times$3 and 1$\times$1 convolutions. In the Bottleneck Transformer \cite{BottleneckTransformers},
the early ResNet blocks are retained as is, but in the last of the blocks, the 3$\times$3 convolution is replaced with a MHSA module.
A schematic of the MHSA block is shown in Figure \ref{fig:bottleneck} (b). 
This MHSA performs a global ({\it all2all}) self-attention over a 2D feature map. The authors of \cite{BottleneckTransformers}  
demonstrated strong performance on the ImageNet classification and COCO instance segmentation tasks.

Self-supervised learning (SSL) is a branch of machine learning that aims to learn useful features in 
data without the use of a supervisory learning signal such as a label. Instead, one
often uses a Siamese-like framework with two identical neural network architectures---termed 
``teacher" and ``student"---that see different augmented views of an input image and try to
match features using Knowledge Distillation \cite{SimCLR}, \cite{BYOL}, \cite{BarlowTwins}, \cite{SimCLR}, \cite{DINO}. Augmentations from the same source
image are a positive pair whereas from different sources form a negative pair. The objective here
is to maximize the similarity between positive pairs and minimize the same for negative pairs \cite{SimCLR}. 
However, only positive pairs are used in many newer self-supervised learning algorithms such as BYOL \cite{BYOL},
Barlow Twins \cite{BarlowTwins}, SimSIAM \cite{SimSiam} and DINO \cite{DINO}. This works by Knowledge Distillation \cite{KnowledgeDistillation}.
Pre-training the network using SSL has the advantage of starting with superior weights vis-\'a-vis random weights, and this
forms the motivation for us to consider self-supervised pre-training.  

While CNN architectures have a proven track record at
low-level feature extraction, they may not be robust enough for determining the local-to-global 
correspondences as they need to be deep enough to have a large receptive field.
Inspired by the success of Bottleneck Transformers \cite{BottleneckTransformers}, in this study we
extend the idea to the problem of 2D human keypoint estimation on the COCO dataset \cite{COCO}.
We use hybrid backbone architectures combining (1) convolutional blocks, (2) MHSA bottleneck block(s) and (3) Transformer
Encoder layer. The early convolutional blocks extract low-level features, whereas the MHSA bottleneck block(s)
use them to extract mid-level features. Finally, the Transformer uses the mid-level features to learn
local-to-global relationships. Our architecture is similar in spirit to a recent paper
called Transpose \cite{Transpose}, however our architecture differs from them in that we use the 
MHSA bottleneck transformer block as a mid-level feature extractor, which they did not consider. 
We test the hypothesis that separating out early, mid and high level features with appropriate attention-based architectures helps in
better learning the local-to-global dependencies, as will be demonstrated in this paper.
The last attention layer of the Transformer Encoder layer acts as a feature aggregator from different sub-regions
of the image and combines them to learn the relationships between them. 

In summary, our contributions are as follows:

\begin{enumerate}   

\item{We use a hybrid architecture that combines Bottleneck Transformers with the vanilla Transformer Encoder to estimate human keypoints by predicting heatmaps. We term our model as BTranspose.}
\item{We consider self-supervised pre-training using DINO \cite{DINO} of our backbone architectures and this is found to improve the prediction accuracy.}
\item{We show attention map dependencies and compare them across different backbone architectures as well as with and without self-supervised pre-training. This provides insights on what exactly the attention layer of the MHSA and the Transformer Encoder focus on in an input image and how it relates to the final keypoint heatmaps predicted.}
\item{With self-supervised pre-training, our best performing BTranspose architecture achieves an AP of 76.4 and an AR of 79.2 on the COCO validation set, which 
is competitive with Transpose \cite{Transpose} on AP with much fewer number of neural network parameters (10.1M for BTranspose versus 17.5M for Transpose).}

\end{enumerate}

\section{Related Work}
\subsection{Human pose estimation}

Deep CNNs have a proven track record at human pose estimation. However, locality and translation equivariance are inductive biases inherent in CNNs \cite{Lenet}, \cite{Alexnet}.
While CNNs are good at extracting low-level features from images, they have to be very deep in order to learn the local-to-global relationships between different
sub-regions of an image. Learning this relationship is vital to the task of human pose estimation as the location of the different human body parts must be learned to preserve 
the human anatomy in the predictions. Many strategies have been proposed such as stacking \cite{Xiao2018}, \cite{Wei2016}, high-resolution representation \cite{Sun2019}
and multi-scale fusion \cite{Pfister2015}, \cite{Newell2016}, \cite{Chu2017}, \cite{Cheng2020} with a good amount of success.
Different neural network architectures have also been considered for human pose estimation: FPN \cite{FPN}, SimpleBaseline \cite{Xiao2018}, Hourglass Network \cite{Newell2016},
HRNet \cite{Sun2019}, RSN \cite{RSN}. Some of these architectures have $>$ 50M neural network parameters, which is hard to train and deploy. In contrast, we focus in this
study on lightweight neural network architectures ($\sim$ 10M parameters or fewer) for human pose estimation so that one can achieve real-time inference speeds.

\subsection{Multi-head self attention}
The basic building block of the vanilla Transformer is the multi-head self attention (MHSA) module \cite{Vaswani2017}.
Since the development of transformers for NLP applications, they have also percolated into computer vision applications.
For instance, DETR \cite{DETR} combines a ResNet-50 and a Transformer Encoder for object detection and reported high AP scores. 
Bottleneck Transformers \cite{BottleneckTransformers} include the multi-head self attention module inside ResNet-like blocks and
have shown good performance for image classification and object 
detection tasks; it forms the basis of the present study.
The Vision Transformer (ViT) \cite{Dosovitskiy2020} extended the idea of MHSA to image patches and has gained a lot of interest in the
computer vision community. DeiT \cite{DeiT} uses a distillation token for the student to learn from the teacher; they also introduce multiple
transformer architectures for computer vision applications: small, medium and large.  
In this study, we extend \cite{Transpose} architecture by using a Bottleneck Transformer in the backbone network and demonstrate robust 
predictions for the problem of 2D human pose estimation.

\subsection{Self-supervised learning}

Self-supervised learning (SSL) is the learning of representations in data without supervised labels and has gained a lot of interest in the computer
vision community recently. BYOL \cite{BYOL} uses two neural networks, referred to as online and target networks.
The online network is trained from an augmented view of an image to predict the representation of the target network of the same image under a different augmented view, using 
the L2 loss function. The target network parameters are a slow-moving average of the online network. 
DINO \cite{DINO} is similar in spirit as BYOL, but uses the cross-entropy loss instead. 
Barlow Twins \cite{BarlowTwins} measures the cross-correlation matrix between the outputs of two identical networks fed with distorted versions of an image, and 
makes it as close as possible to the identity matrix. This enforces the embedding vectors of augmented images to be similar, and concomittantly minimizes
the redundancy between the components of these vectors. Recently, \cite{Data2vec} use the same SSL framework for speech, NLP and computer vision, by 
predicting latent representations of the full input data based on a masked view of the input in a self-distillation setup with standard Transformer architectures.  
Inspired by the success of these different SSL methods for multiple applications, in this study we use the DINO method to pre-train the backbone architectures
in a SSL fashion on ImageNet images (i.e., without class labels) and transfer the learning to 2D human pose estimation.
We will demonstrate that this SSL pre-training consistently augments the final prediction AP values.

\subsection{Explainability of neural networks}

Explainability is essentially an effort to understand what the model learns from data to make predictions. 
This is a key component in Deep Learning as otherwise neural network predictions are essentially black box in nature lacking an understanding of what the model infers from data.
Explainability and interpretability of a neural network is ascertaining what parts of an input are the most relevant to make predictions \cite{Samek2019}.
\cite{Li2014} undertake gradient descent in the input to find which patterns maximize a given unit. 
For image classification task, \cite{Simonyan2013} compute a class saliency map, specific to a given image and class and show that such maps can be used
for weakly supervised object segmentation using CNN architectures. They also establish the connection between gradient-based visualization methods in CNNs 
and deconvolutional networks. 
\cite{Zeiler2014} introduced a novel visualization technique to provide insights into the function of intermediate feature layers in CNNs. 
Transpose \cite{Transpose} also present dependency areas for keypoints for the 2D human pose estimation problem using 
Activation Maximization \cite{Simonyan2013}. Explainability boils down to a sensitivity analysis problem of estimating which 
region of an input maximizes the activation of a given neuron. Similar to \cite{Transpose}, we also undertake such dependency area estimation, but
the main difference being that we make this dependency analysis both at the mid (MHSA block) and high levels (Transformer Encoder) whereas they do so only at the high level.
We will demonstrate how this mid-level explainability is useful to better understand what the network learns.


\begin{figure*}[!h]
\begin{center}
  \includegraphics[width=1.0\linewidth]{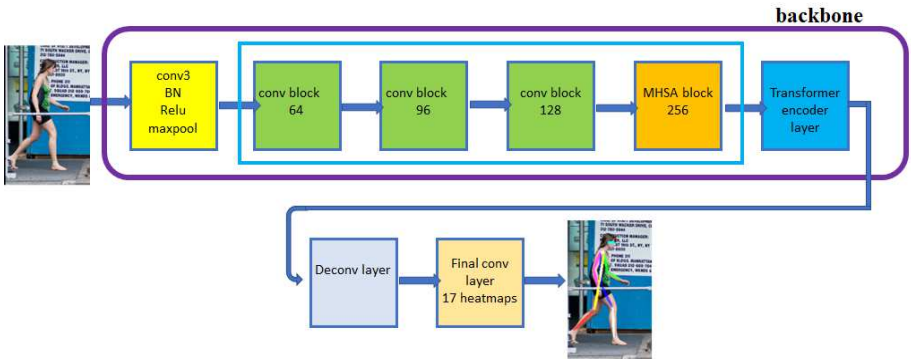}
\end{center}
   \caption{The BTranspose architecture: shown here is the C3A1 architecture used in our paper. The purple rectangle is the backbone used. The naming convention for the different architectures is bsed on the number of CNN and MHSA blocks in the blue rectangle inside the backbone. Finally, the output of the Transformer encoder is fed into a Deconvolutional layer followed by a 1$\times$1 convolutional layer to output 17 heatmaps.}
\label{fig:btranspose}
\end{figure*}

\section{Method}

The primary objective of this study is to make 2D human keypoint predictions from input images, and also explain global dependencies between the keypoint
predictions. Specifically, we explore different backbone architectures by combining Bottleneck Transformers \cite{BottleneckTransformers} 
and vanilla Transformer encoder layer \cite{Vaswani2017}. The backbone architectures are pre-trained using Self-Supervised Learning (SSL) and we
will demonstrate that this strategy improves the prediction accuracy. Most of the architectures considered in this study are lightweight and have $\sim$ 10M parameters. 

\subsection{Architecture}

Our neural network architecture consists of a backbone for feature extraction, followed by a head to predict 17 keypoint heatmaps.
A schematic of one of our architectures is presented in Figure \ref{fig:btranspose}. 
We will now describe our neural network architectures, indicating the color used for each module in Figure \ref{fig:btranspose}. 
The purple rectangle in Figure \ref{fig:btranspose} represents the backbone, which is pre-trained using self-supervised learning (SSL), and is now described in detail.

{\bf Backbone}
The backbone consists of three components: (1) early CNN block; (2) a Bottleneck Transformer \cite{BottleneckTransformers}; (3) a vanilla Transformer encoder layer \cite{Vaswani2017}.
The input image first goes through an early convolutional block (yellow in Figure \ref{fig:btranspose})
that consists of 7$\times$7 convolutional kernel, Batch Normalization and max pooling to downsample, identical to a ResNet-50 \cite{Resnet}.
The second module in the backbone is the Bottleneck Transformer (BT) \cite{BottleneckTransformers} with consists of
either 2 or 3 convolutional ResNet block groups \cite{Resnet}, followed by 1 or 2 multi-head self-attention (MHSA) block groups (see Figure \ref{fig:bottleneck}). 
The convolutional ResNet blocks consist of 3$\times$3 convolutions sandwiched between
1$\times$1 convolutions (green in Figure \ref{fig:btranspose}), albeit the number of feature maps used in this study for these convolutional blocks are fewer than the standard ResNet.

This is followed by the MHSA blocks \cite{BottleneckTransformers} (refer to Figure \ref{fig:bottleneck}) with the 
MHSA layer sandwiched between 1$\times$1 convolutional layers (orange in Figure \ref{fig:btranspose}).
The naming convention we will use in this paper is the number of such blocks in the BT. 
For example, architectures with 2 CNN blocks and 2 MHSA blocks are termed C2A2; with 3 CNN blocks and 1 MHSA block are termed C3A1, etc.
The blue rectangle in Figure \ref{fig:btranspose} denotes the subset of the backbone that gives rise to the naming convention used: 
for instance, since we use 3 conv blocks (green) and 1 MHSA block (orange), the architecture shown in the figure is C3A1.
 
The last module in the backbone is a standard Transformer encoder layer (blue in Figure \ref{fig:btranspose}) to capture the long-range
dependencies from the sequences using the standard query-key-value attention modules \cite{Vaswani2017}. 
If the output of the MHSA block is $\mathbb{R}^{(d_1 \times H \times W)}$, it is first passed through a 1$\times$1 convolution to obtain a feature dimension $d$ and is
then flattened before feeding into the Transformer encoder. The dimension of the features input to the Transformer encoder is $\mathbb{R}^{(L \times d)}$, where $L = H \times W$.
This goes through $N$ attention layers and feed-forward networks (FFNs), similar to the vanilla Transformer \cite{Vaswani2017}.     
Note that our architecture is somewhat similar to \cite{Transpose}, but they used vanilla CNN blocks followed by the Transformer encoder, whereas we use 
BT followed by the vanilla Transformer encoder. 
The motivation for choosing our architectures is that the CNN blocks in BT capture the low-level features, the MHSA blocks
use this information to capture mid-level features using attention, and the Transformer encoder will then use these mid-features to learn the global relationships. 
Thus, our architecture uses self-attention both at the mid and high levels. 

{\bf Head} A head is attached to the backbone. This head consists of a deconvolution layer (grey in Figure \ref{fig:btranspose}) 
to increase the size of the feature map from 32$\times$24 to 64$\times$48, and a simple 1$\times$1 convolutional layer (cream color in Figure \ref{fig:btranspose})
to output 17 human keypoint heatmaps of size 64$\times$48. 

\subsection{DINO self-supervised pretraining}

The self-supervised pre-training of the backbone is undertaken using the DINO \cite{DINO} method, which uses self distillation from the 
teacher to student networks. Augmented views of the input images are fed into the teacher and student networks, respectively.  
The output features of the teacher and student are enforced to be similar with the cross-entropy loss, which is back-propagated into the 
student network. The teacher network follows the student network by means of an Exponential Moving Average (EMA) update rule, i.e., a Momentum Encoder \cite{He2020}.
In addition, a centering operation is done to avoid collapse \cite{DINO}. 

\section{Experiments}

{\bf Dataset}. For the Self-Supervised pre-training of the backbone architectures, we use ImageNet (images only, no labels). 
For the 2D human pose estimation downstream task, all our training and inference is on the COCO \cite{COCO} dataset. COCO consists of 200k images in the wild, with $\sim$ 250k human instances.
Train2017 has 57k images and 150k person instances. Val2017 has 5k images. The evaluation metric is based on Object Keypoint Similarity (OKS) \cite{COCO} and 
we compare Average Precision (AP) and Average Recall (AR) for the different models on the COCO validation set.

{\bf Training details}. The SSL pre-training strategy follows \cite{DINO} and uses 300 epochs. 
For the 2D human pose estimation, we follow the top-down human pose estimation paradigm where a pre-trained object detector
obtains the bounding box around humans in an image, which is then extracted and
fed into the pose estimator. The training set consists of single person crops from the dataset. All input images are resized to 256 $\times$ 192. 
The same data augmentation and training strategies from \cite{Sun2019}, \cite{Transpose} are used in this study. We use a total of 230 training epochs for 
the 2D human pose estimation problem, and employ the L2 loss function.

{\bf Architecture details}. The architectures we use comprise of either 3 or 4 block groups \cite{BottleneckTransformers}
with the number of feature maps being [64, 96, 128, 256] for the block groups (in the same sequence),
which is fewer than the original BT used in \cite{BottleneckTransformers} or the ResNet-50. The number of blocks used in each group is
[3, 4, 6, 3], which is the same as the ResNet-50 and BT.
We vary the number of attention heads in the MHSA block and this takes the value 4 or 8 in this study.  
The Transformer encoder uses a transformer dimension of 256 and a feed-forward dimension of 1024. The number
of transformer layers is $N$ = 4 (except in one ablation study where $N$ = 6), the number of heads in the Transformer is 8.
We use relative position embedding for height and width for the MHSA block, consistent with \cite{BottleneckTransformers}.
For the Transformer encoder, we use learnable position embeddings \cite{Vaswani2017}.

{\bf Naming convention}. As aforementioned, the naming convention used in this paper (e.g., C2A1, C2A2, C3A1) is based on the number of convolutional and MHSA blocks in the BT part of
the backbone (inside the blue rectangle shown in Figure \ref{fig:btranspose}). 
When the number of attention heads in the MHSA block is 4, we add a ``(4)" to the model name; and similarly a ``(8)" when 8 heads are used in the MHSA block (note: the number
of attention heads in the Transformer encoder is 8 for all the cases considered in this paper).
In addition, when we pre-train with DINO, we add a suffix ``Dino" to the model name; if the suffix ``Dino" is not used, it is a model
trained on the 2D pose estimation task with random weights as initialization.


\subsection{Results on 2D human pose estimation}

We compare BTranspose with SimpleBaseline \cite{Xiao2018}, HRNet \cite{Sun2019} and Transpose \cite{Transpose} in Table \ref{table1} on the
COCO validation set.
Specifically, C3A1(4)-Dino is the best performing BTranspose model with an AP of 76.4 and AR of 79.2, and so the comparison is done with this.
C3A1(4)-Dino outperforms all the other models shown in Table \ref{table1} on AP and comes close to Transpose-H-A4/6 on AR.
Note that C3A1(4)-Dino only uses 10.1M neural network parameters, whereas Transpose-H-A4/6 use 17.5M.
C3A1(4)-Dino also outperforms Transpose-H-S which has number of parameters comparable to BTranspose.
Thus, with only 10.1M parameters, BTranspose is competitive with other pose estimation models. 
Furthermore, at inference C3A1(4)-Dino has an FPS of 45, which is also comparable with other pose estimation models such as \cite{Transpose}.  

\begin{table}
\begin{center}
\begin{tabular}{|c|c|c|c|}
\hline
{\bf Model} & {\bf AP} & {\bf AR} & {\bf \#Params} \\
\hline
SimpleBaseline-Res50 \cite{Xiao2018} & 70.4 & 76.3 & 34.0M \\
SimpleBaseline-Res101 \cite{Xiao2018} & 71.4 & 76.3 & 53.0M \\
SimpleBaseline-Res152 \cite{Xiao2018} & 72.0 & 77.8 & 68.6M \\
\hline
HRNet-W32 \cite{Sun2019} & 74.4 & 79.8 & 28.5M \\
HRNet-W48 \cite{Sun2019} & 75.1 & 80.4 & 63.6M \\
\hline
Transpose-R-A3 \cite{Transpose} & 71.7 & 77.1 & 5.2M \\
Transpose-R-A4 \cite{Transpose} & 72.6 & 78.0 & 6.0M \\
Transpose-H-S \cite{Transpose} & 74.2 & 78.0 & 8.0M \\
Transpose-H-A4 \cite{Transpose} & 75.3 & 80.3 & 17.3M \\
Transpose-H-A6 \cite{Transpose} & 75.8 & {\bf 80.8} & 17.5M \\
\hline
C3A1(4)-Dino (this study) & {\bf 76.4} & 79.2 & 10.1M \\
\hline
\end{tabular}
\end{center}
\caption{Comparison of AP and AR on the COCO validation set for different models. C3A1(4)-Dino achieves competitive results with fewer parameters.}
\label{table1}
\end{table}


\subsection{Ablation studies}

{\bf SSL pre-training versus random weight initialization}. Self-supervised learning (SSL) used as pre-training has a proven track record 
of learning useful image representations from unlabeled data. To this end, we have also pre-trained our models on
ImageNet (images only, no lables) using the DINO \cite{DINO} method. We compare our pose estimation models with (1)
random weight initialization and (2) DINO pre-training on ImageNet. With pre-training, the neural network learns useful
low level features such as basic shapes and edges and how to aggregate information from them for the downstream pose
estimation task. Table \ref{table2} compares several BTranspose models with random initialization and
DINO pre-training.\footnote{training C2A2(4) went awry, probably due to poor weight initialization.}
All three BTranspose models C2A1(4)-Dino, C2A2(4)-Dino, C3A1(4)-Dino outperform when pre-trained using DINO, with +1.3 AP and +1.4 AR observed for
C3A1(4)-Dino compared to its random initialization counterpart C3A1(4). Pre-training using SSL obviously improves performance, even though the 
pre-training was done on a completely different dataset (i.e., ImageNet), as basic features such as shapes and edges can still be learned which 
is useful for the downstream pose estimation task. Furthermore, C3A1(4)-Dino outperforms both C2A1(4)-Dino and C2A2(4)-Dino, which suggests that 
3 convolutional blocks help in learning better low-level features, before we proceed to processing the mid-level features in the MHSA block.    

\begin{table}
\begin{center}
\begin{tabular}{|c|c|c|c|}
\hline
{\bf Model} & {\bf AP} & {\bf AR} & {\bf \#Params} \\
\hline
\multicolumn{4}{|c|}{random initialization} \\
\hline 
C2A1(4) & 74.2 & 76.9 & 6.8M \\
C2A2(4) & 59.0 & 62.6 & 9.55M \\
C3A1(4) & 75.1 & 77.8 & 10.1M \\
\hline
\multicolumn{4}{|c|}{self-supervised pre-training}  \\
\hline
C2A1(4)-Dino & 75.2 & 78.0 & 6.8M \\
C2A2(4)-Dino & 75.0 & 77.8 & 9.55M \\
C3A1(4)-Dino & 76.4 & 79.2 & 10.1M \\
\hline
\end{tabular}
\end{center}
\caption{Random weight initialization versus self-supervised pre-training on ImageNet using DINO \cite{DINO}.}
\label{table2}
\end{table}

{\bf 4 versus 8 heads in the MHSA block}. For the MHSA block of the Bottleneck Transformer used in the backbone,
we considered either 4 or 8 self-attention heads (note: the Transformer encoder always uses 8 heads for all models; the analysis here is for the MHSA block only).
The MHSA block is responsible for learning mid-level features using attention, which are then passed on to the Transformer encoder. 
The results are compared in Table \ref{table3} for C3A1 architecture for the bottleneck in the backbone. 
With random weight initialization, 8 heads outperforms 4 heads, with +0.2 and +0.3 on AP and AR observed, respectively. 
However, with SSL pre-training, 4 heads has better performance than 8: +0.4 on AP and +0.3 on AR. 
We believe that more heads help training when the networks are trained from scratch with random weight initialization.
However, when SSL pre-training is used, the networks already learn useful image features and in this setting fewer heads in the MHSA block suffices.

\begin{table}[!htb]
    \begin{minipage}{.5\linewidth}
      \begin{tabular}{|c|c|c|}
      \hline
      {\bf Model} & {\bf AP} & {\bf AR}  \\
      \hline
      \multicolumn{3}{|c|}{random initialization} \\
      \hline 
      C3A1(4) & 75.1 & 77.8  \\
      C3A1(8) & 75.3 & 78.1  \\
      \hline
      \multicolumn{3}{|c|}{self-supervised pre-training}  \\
      \hline
      C3A1(4)-Dino & 76.4 & 79.2  \\
      C3A1(8)-Dino & 76.0 & 78.9  \\
      \hline
      \end{tabular}
      \caption{4 versus 8 attention heads in the MHSA block. All models have 10.1M \#Params.}
      \label{table3}
    \end{minipage}%
    \begin{minipage}{.5\linewidth}
        \begin{tabular}{|c|c|c|c|}
        \hline
        {\bf Model} & {\bf AP} & {\bf AR} & {\bf \#Params}  \\
        \hline
        C3A1(4)-Dino & 76.4 & 79.2 & 10.1M  \\
        C3A1(4)-Dino-N6 & 76.3 & 79.1 & 11.67M \\
        C3A1(4)-Dino-Large & 76.7 & 79.3 & 23.82M \\
        \hline
        \end{tabular}
        \caption{Other architecture variants.}
        \label{table3b}
    \end{minipage} 
\end{table}

{\bf Number of transformer encoder layers}. We consider another case where the number of transformer encoder layers is varied. 
For all the cases considered thus far, we used N = 4 layers; we consider one case with N = 6 layers (called C3A1(4)-Dino-N6), with everything else
being the same. The results on the COCO validation set are presented in Table \ref{table3b}. Compared to the baseline
C3A1(4)-Dino case, AP and AR are -0.1 and -0.1, respectively. Thus, adding extra transformer encoder layers has no performance boost.

{\bf Larger network architecture}. The number of feature maps used in the CNN and MHSA blocks for all the cases considered thus far is 
[64, 96, 128, 256]. We consider a larger network architecture with [64, 128, 256, 512] feature maps for the CNN and MHSA block groups---we name it C3A1(4)-Dino-Large---
whose backbone is the same as the original architecture used in \cite{BottleneckTransformers}. As before, this network is also pre-trained using DINO \cite{DINO} and the 
results on the COCO validation set are presented in Table \ref{table3b}. AP and AR for C3A1(4)-Dino-Large are
+0.3 and +0.1 compared to C3A1(4)-Dino, which is an incremental improvement, albeit with 2X number of parameters. 

{\bf Optimal transport loss function}. We also consider one training with the Sinkhorn loss function used in optimal transport \cite{Cuturi2013}
for the downstream pose estimation task, with the entropy regulation term $\epsilon$ = 0.05 and number of training iterations $n$ = 3.
On the COCO validation set the AP is 76.3 and AR is 79.0, which is -0.1 and -0.2 compared to C3A1(4)-Dino. 
Thus, with the Sinkhorn loss BTranspose slightly underperforms. We believe the reason could be the two hyperparameters $\epsilon$ and $n$
may need to be calibrated, which is problem specific. 

\begin{figure*}[!h]
\centering%
(a)\includegraphics[width=10cm]{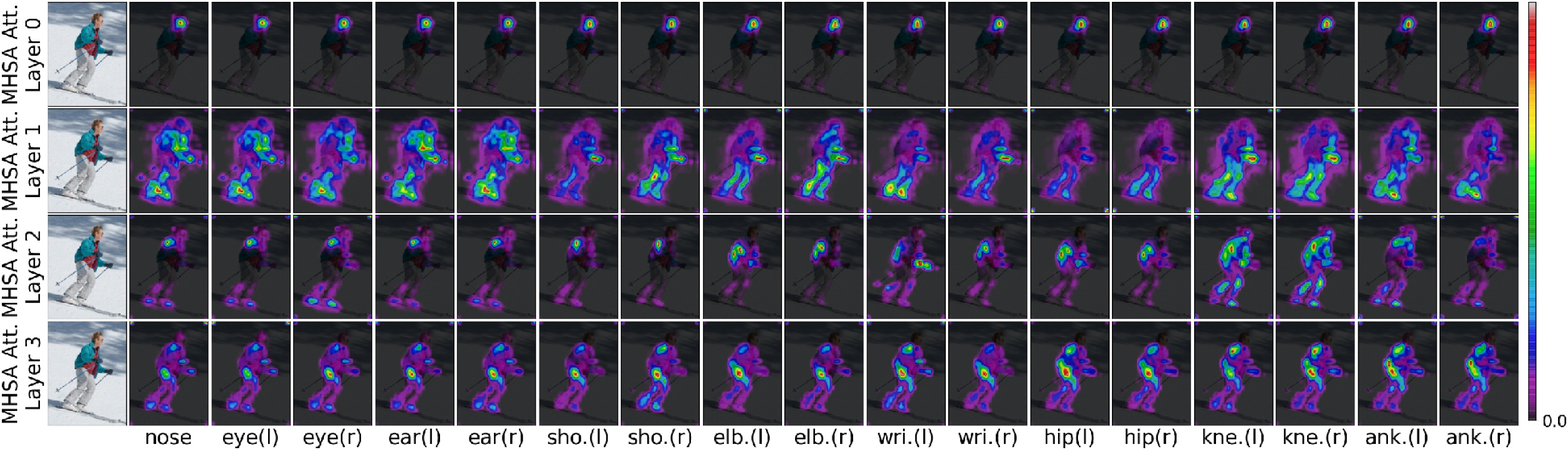} \\
(b)\includegraphics[width=10cm]{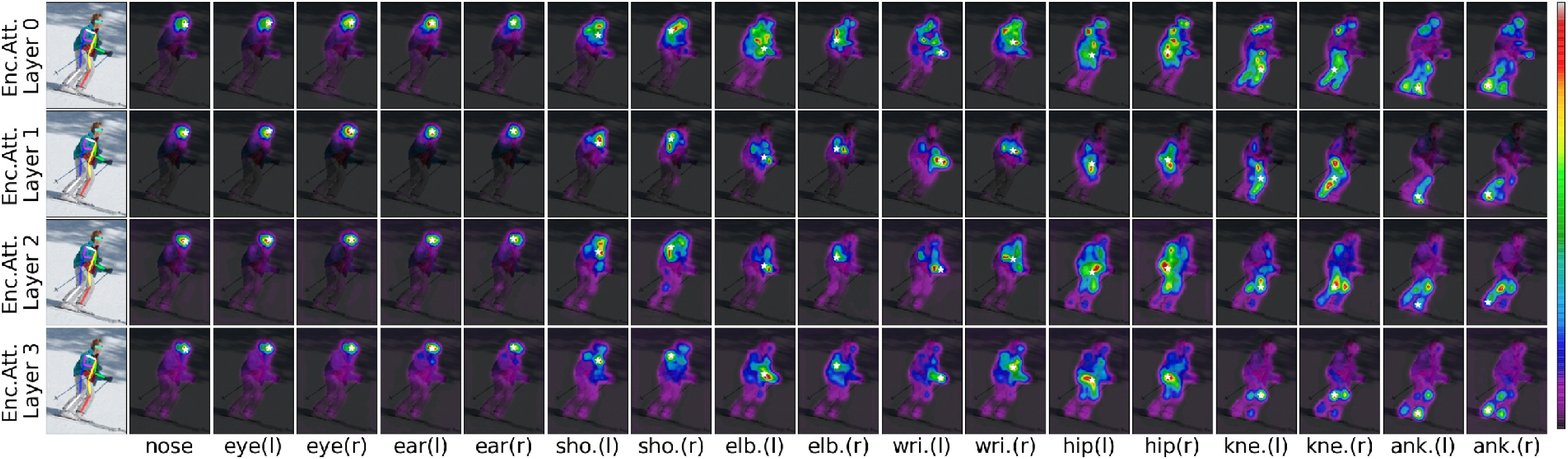} \\
(c)\includegraphics[width=10cm]{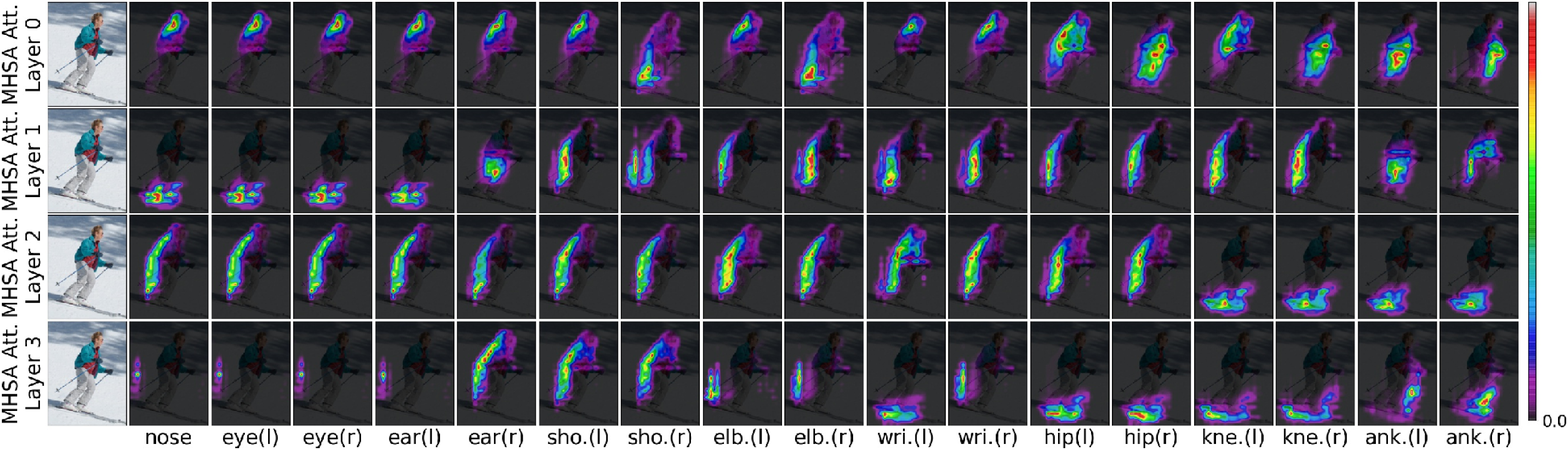}  \\
(d)\includegraphics[width=10cm]{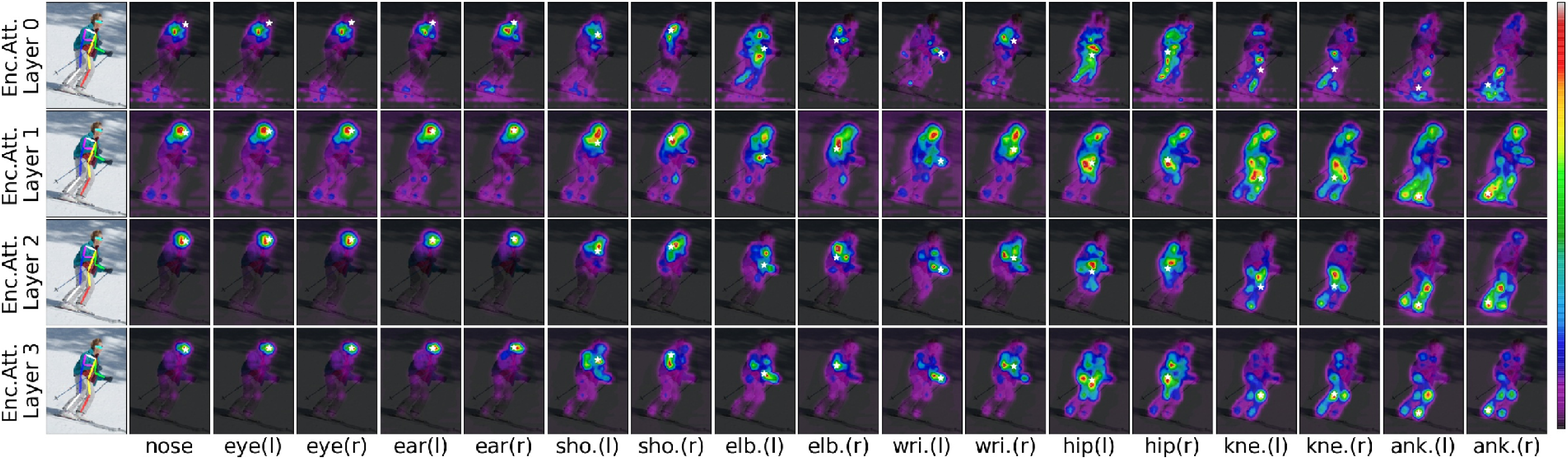} \\ 
\caption{Dependency areas for the predicted keypoints using models: (a \& b) C3A1(4)-Dino; (c \& d) C3A1(4). The dependency area is computed at the MHSA block in (a \& c) and at 
the Transformer encoder layer in (b \& d). The predicted keypoints and the skeleton reconstruction is also shown in the first column of each image.}
\label{fig:da1}%
\end{figure*}

\subsection{Dependency areas}

We undertake a qualitative analysis to understand how the network makes
the final predictions by aggregating information. To this end, we visualize dependency areas from attention maps
using the final predicted keypoint as the query location, similar to \cite{Transpose}. But the analysis
differs from \cite{Transpose} in that they have attention layers only in the Transformer encoder, whereas in our study
we have 2 parts of the backbone network that have attention layers: the MHSA block in the bottleneck transformer and in the 
final Transformer Encoder (TE) layer. The dependency areas in the MHSA block provides insights on the role of attention at the 
mid-level features; whereas the same at the TE layer provides insights on the role of attention at the high-level features. 

{\bf MHSA block versus Transformer encoder} We present the dependency areas for an input image comparing 
C3A1(4)-Dino and C3A1(4) in Figure \ref{fig:da1}. It is interesting to note that at the MHSA level, the network has already learned to
focus on the human anatomy/shape, i.e., a rough segmented region of the human,
with many potential keypoint regions having high scores (yellow/red patches). 
However, at the MHSA block (Figure \ref{fig:da1} a \& c), no strong correlation is evident between the 
final keypoint predicted (white star in the figure) and the region where the MHSA block focuses on in the input image. 
In the Transformer encoder attention dependency areas (Figure \ref{fig:da1} b \& d), 
we see high values in the vicinity of the predicted keypoints. 
This suggests that at the MHSA block, the network has learned to aggregate low-level information to construct mid-level features and 
infer the basic human anatomy/shape, but has not yet learned which part of the human anatomy corresponds to the final keypoint of interst.
In contrast, at the Transformer encoder level, the network aggregates the mid-level understanding to learn local-to-global relationships, and so 
arrives at the final predicted keypoint.

\begin{figure*}[!h]
\centering%
(a)\includegraphics[width=10cm]{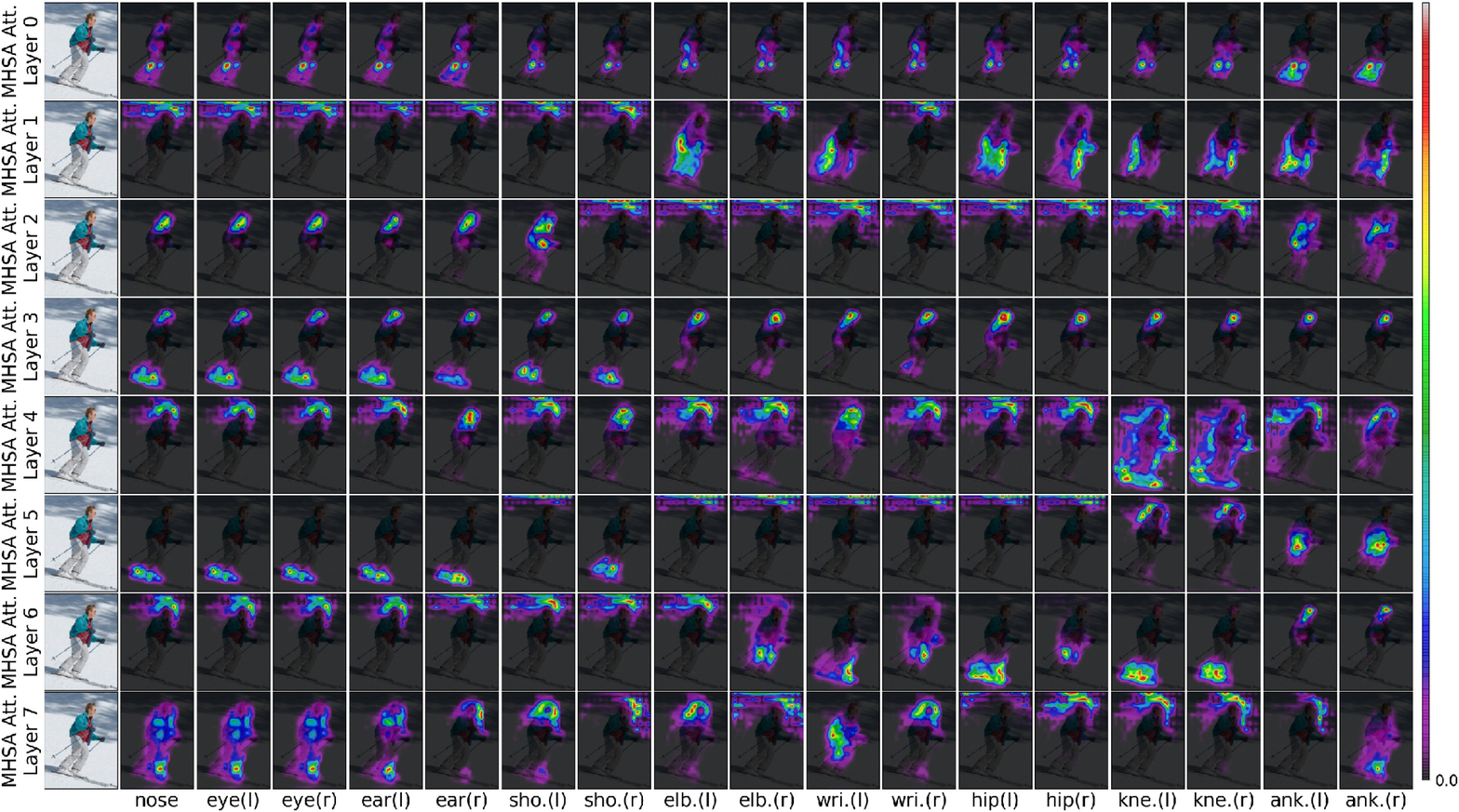} \\
(b)\includegraphics[width=10cm]{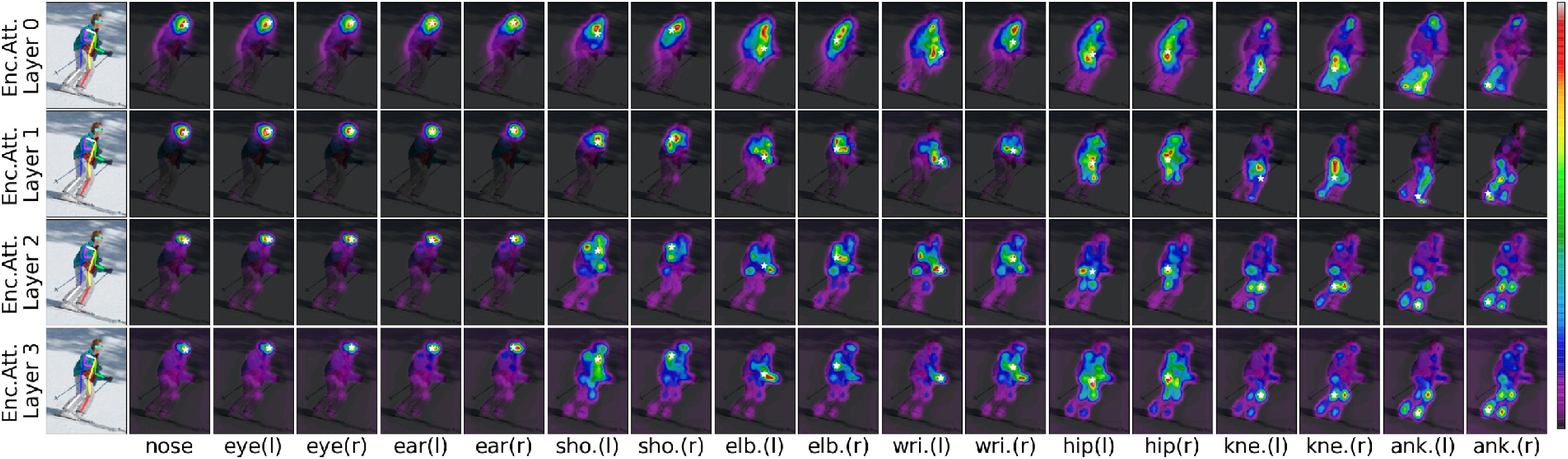} \\
\caption{Dependency areas for the predicted keypoints using C3A1(8)-Dino for the (a) MHSA block and (b) Transformer encoder layer.}
\label{fig:da2}%
\end{figure*}

{\bf Dependency area due to self-supervised pre-training.} Comparing C3A1(4)-Dino and C3A1(4) in Figure \ref{fig:da1}, it is evident that the 
dependency areas are different. At the MHSA level, the dependency areas look relatively smaller for C3A1(4)-Dino than C3A1(4), so 
pre-training potentially helps at the MHSA level as the early layers have learned low-level features better. At the Transformer encoder level, the 
differences are subtle, suggesting that even though there may be multiple pathways by which the mid-level features are extracted in the MHSA block, at the 
Transformer encoder layer they still get aggregated to form mid-to-high relationships to arrive at the final keypoint. More examples are presented in the
supplementary material.

{\bf 4 versus 8 attention heads in the MHSA block.} Next, we consider C3A1(8)-Dino and present the dependency areas for the predicted keypoints 
in Figure \ref{fig:da2}. Here too the network has learned the basic human anatomy (i.e., a segmented sub-region of the human) at the MHSA block using attention (i.e., the mid-level), and 
at the Transformer Encoder level it has learned to aggregate the mid-level information to infer the global relationships and predict the final keypoint. 
Comparing Figures \ref{fig:da1} and \ref{fig:da2}, we observe that at the MHSA block 8 heads seems to result in a few dependency areas 
that are partially outside the human sub-region, suggesting that 8 heads can confuse the network to some extent at the mid-level features. 
However, at the Transformer Encoder level, the differenes are much less pronounced, suggesting that at the high-level features the Transformer Encoder has overcome some of the
confusion that exists at the mid-level. More examples are presented in the supplementary material. 

\begin{figure*}[h]
\centering%
(a) \includegraphics[width=5.5cm]{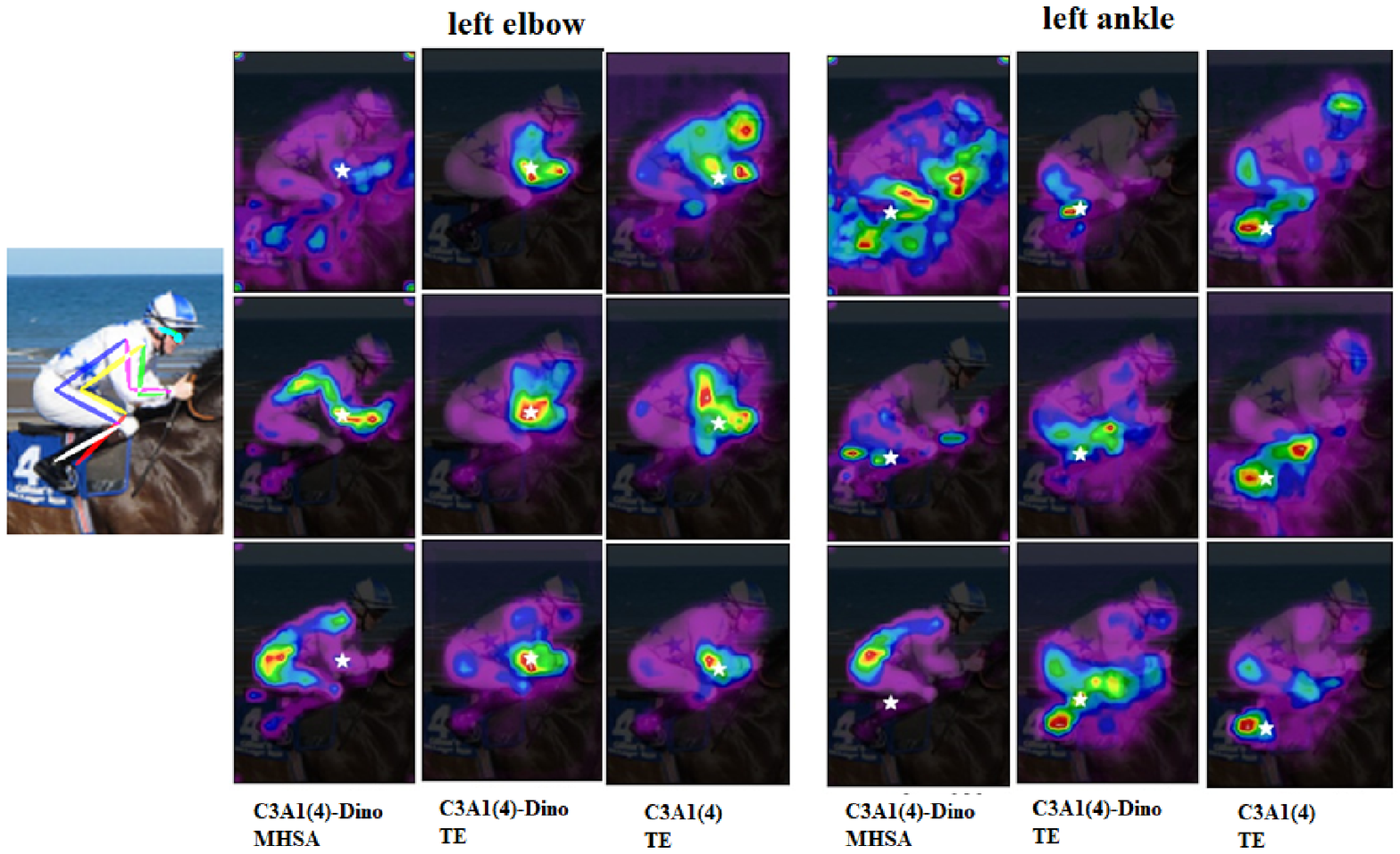} (b) \includegraphics[width=5.5cm]{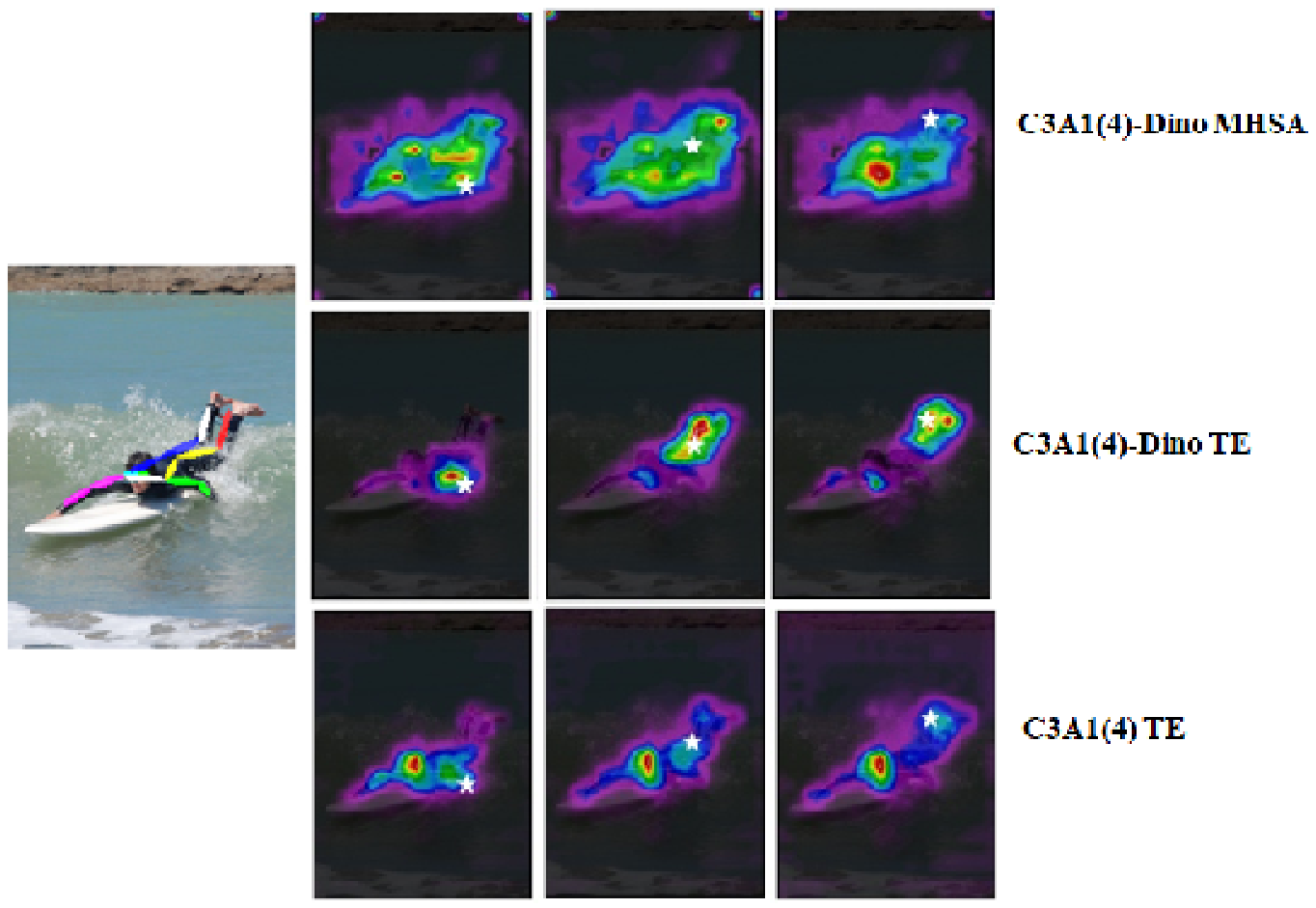}
\caption{Dependency areas for the (a) horse rider and the (b) surfer. The white star denotes the predicted keypoint location.}
\label{fig:da21}%
\end{figure*}

{\bf Special Scenarios.} Next, we consider some special cases and compare the dependency areas. In Figure \ref{fig:da21} (a) we consider a horse rider with the left half of the human 
occluded. At the MHSA block, the network has already learned the shape of the human, although some 
regions of the horse are also attended to. 
Comparing the dependency areas at the TE layer for the left elbow and left ankle (i.e., two occluded body parts) 
we find the red patches (i.e., peak magnitudes) to be closer to the predicted keypoint location for C3A1(4)-Dino than for C3A1(4). 
We consider the input image of a surfer in Figure \ref{fig:da21} (b). This is interesting as the human in the image is nearly horizontal, whereas most of the images
in the training are of vertically-oriented humans. At the MHSA block, the network has learned to focus on the sub-region of interest, albeit with  
no correlation with the query keypoint. At the TE layer, the peak magnitudes are closer to the query keypoint for C3A1(4)-Dino than for C3A1(4)
suggesting that SSL pre-training helps in the network being more confident in its predictions.
In another special case, we consider two humans close to each other in the input image (see supplementary material). Here, the 
MHSA block is unable to distinguish between the two humans, but has learned to segment the sub-region as a large patch.
The TE layer then aggregates this information to make the distinction between the two humans close to each other. 

\section{Conclusion}

For the task of 2D human pose estimation, we explored a model---BTranspose---by combining 
Bottleneck Transformers with the vanilla Transformer Encoder (TE). We have attention layers in two regions of the backbone:
the MHSA block and in the Transformer encoder. Attention in the MHSA block focuses on the mid-level features, whereas attention in the  
TE aggregates this information to learn high-level features.
Self-supervised pre-training of the backbone is found to improve the performance of the models for all the architectures considered in this study.
The model is lightweight and with only 10.1M parameters is able to compete with other state-of-the-art approaches on the COCO dataset. 
Analyzing dependency areas reveals that the MHSA block learns the basic human anatomy/shape, but has not yet learned which 
region of the human anatomy to focus on for the queried keypoint. 
However, this mid-level information is then used by the Transformer encoder to learn the mid-to-high relationships and arrive at the final keypoint.
Analyzing some special scenarios, we find that the model pre-trained with SSL is slightly more confident in knowing where to focus on in the input image.
This is particularly useful for images with occluded human body parts, as well as when there are two humans very close to each other. 
Overall, BTranspose is competitive with other 2D pose estimation models, is lightweight and provides insights into how the model aggregates information at
both the mid and high-levels.

\bibliographystyle{splncs}
\bibliography{kbbib}
\newpage

\section*{A. Image and feature map dimensions for C3A1 architecture}

The image and feature map dimensions at various stages of the C3A1 architecture is shown in Table \ref{table4}. The batch
dimension is not presented. 
The input image is RGB and of size 256 $\times$ 192.
The model predicts 17 keypoint heatmaps, and so the output is 17 $\times$ 64 $\times$ 48.

\begin{table}
\begin{center}
\begin{tabular}{|c|c|}
\hline
{\bf Layer} & {\bf Dimension} \\
\hline
Input image & 3 $\times$ 256 $\times$ 192  \\
Early conv layer & 64 $\times$ 64 $\times$ 48  \\
First conv block & 256 $\times$ 64 $\times$ 48  \\
Second conv block & 384 $\times$ 32 $\times$ 24  \\
Third conv block & 512 $\times$ 32 $\times$ 24  \\
MHSA block & 1024 $\times$ 32 $\times$ 24  \\
1 $\times$ 1 conv  & 256 $\times$ 32 $\times$ 24 \\
Flatten and reshape  & 768 $\times$ 256 \\
Transformer encoder & 768 $\times$ 256 \\
Reshaped output of backbone & 256 $\times$ 32 $\times$ 24 \\
Deconv layer & 256 $\times$ 64 $\times$ 48 \\
Final 1 $\times$ 1 conv & 17 $\times$ 64 $\times$ 48 \\ 
\hline
\end{tabular}
\end{center}
\caption{Image and feature map dimensions for C3A1 architecture at different depths of the network.}
\label{table4}
\end{table}

\section*{B. BTranspose training details}

The training hyperparameters are shown in Table \ref{table5}.

\begin{table}
\begin{center}
\begin{tabular}{|c|c|}
\hline
{\bf Layer} & {\bf Dimension} \\
\hline
Image size &  3 $\times$ 256 $\times$ 192 \\
Heat map size & 64 $\times$ 48 \\ 
Batch size & 16  \\
Number of epochs & 230  \\
Learning rate & 1e-4 to 1e-5 \\
Learning rate steps & 100, 150, 200, 220 \\
Learning rate factor & 0.25 \\
Optimizer & Adam \\
Momentum & 0.9 \\
$\gamma_{1}$ & 0.99 \\ 
$\gamma_{2}$ & 0 \\
GPU used & Nvidia RTX 6000 \\
\hline
\end{tabular}
\end{center}
\caption{Training hyperparameters for the pose estimation task.}
\label{table5}
\end{table}

\section*{C. Additional dependency area visualizations}

\begin{figure*}[h]
\centering%
(a)\includegraphics[width=12cm]{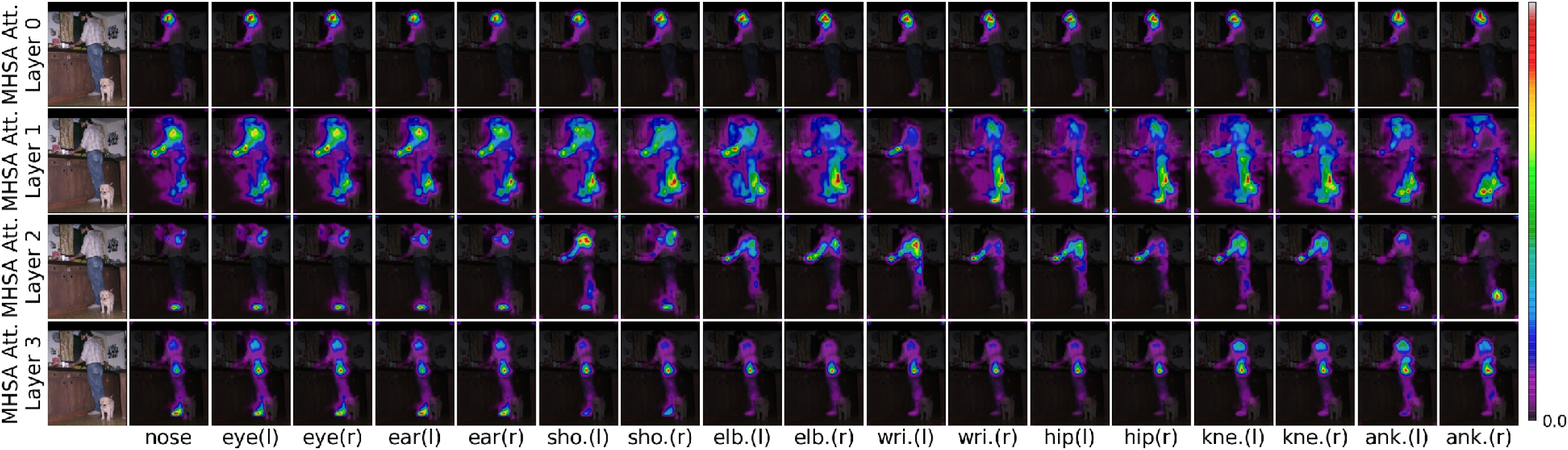} \\
(b)\includegraphics[width=12cm]{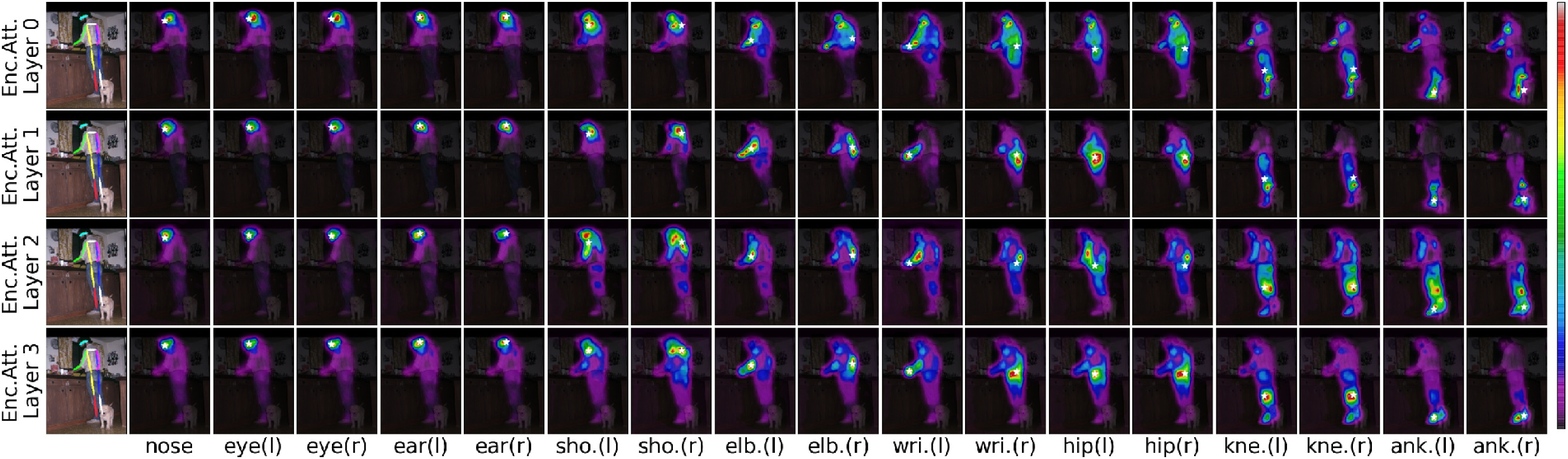} \\
(c)\includegraphics[width=12cm]{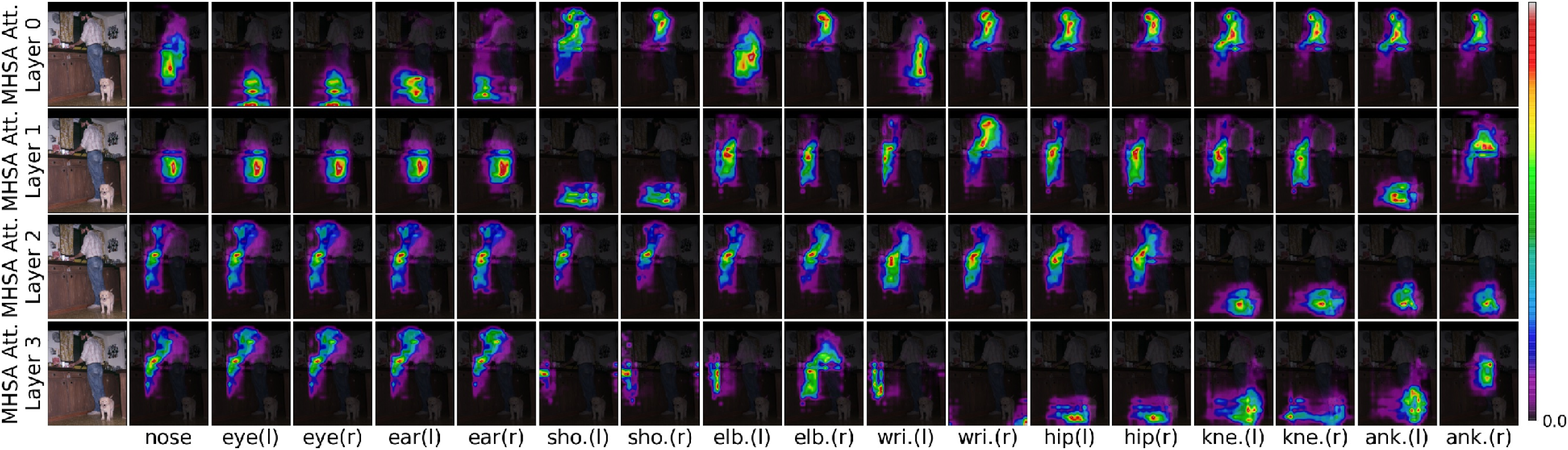}  \\
(d)\includegraphics[width=12cm]{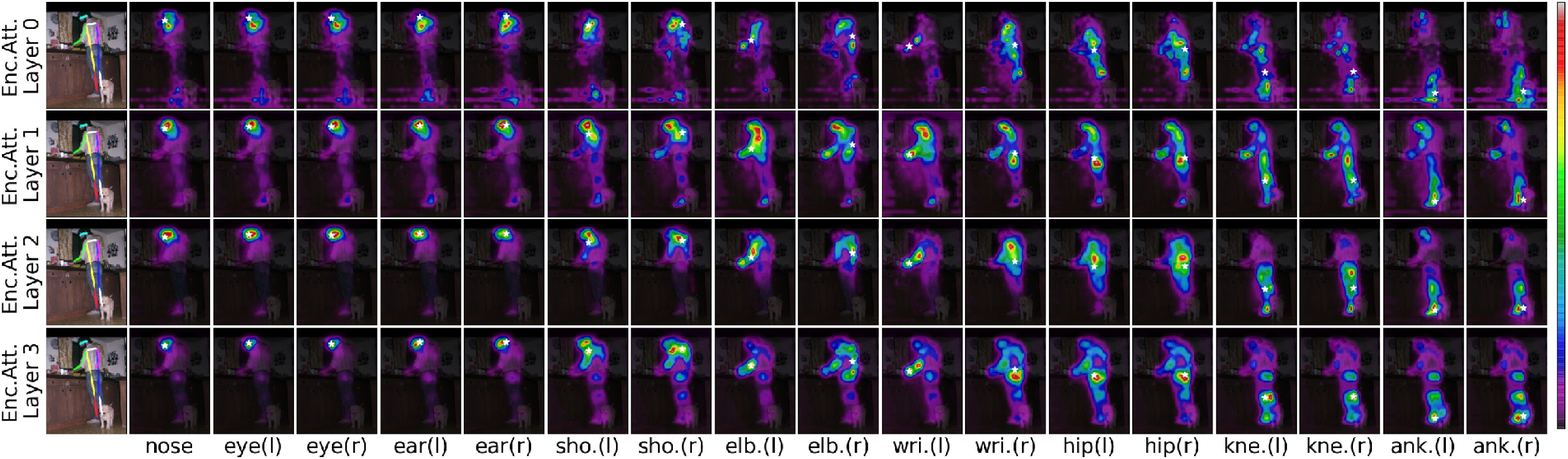} \\ 
\caption{Dependency areas for the predicted keypoints for another input image using models: (a \& b) C3A1(4)-Dino; (c \& d) C3A1(4). The dependency area is computed at the MHSA block in (a \& c) and at the Transformer encoder layer in (b \& d).}
\label{fig:da11}%
\end{figure*}

\begin{figure*}[h]
\centering%
(a)\includegraphics[width=12cm]{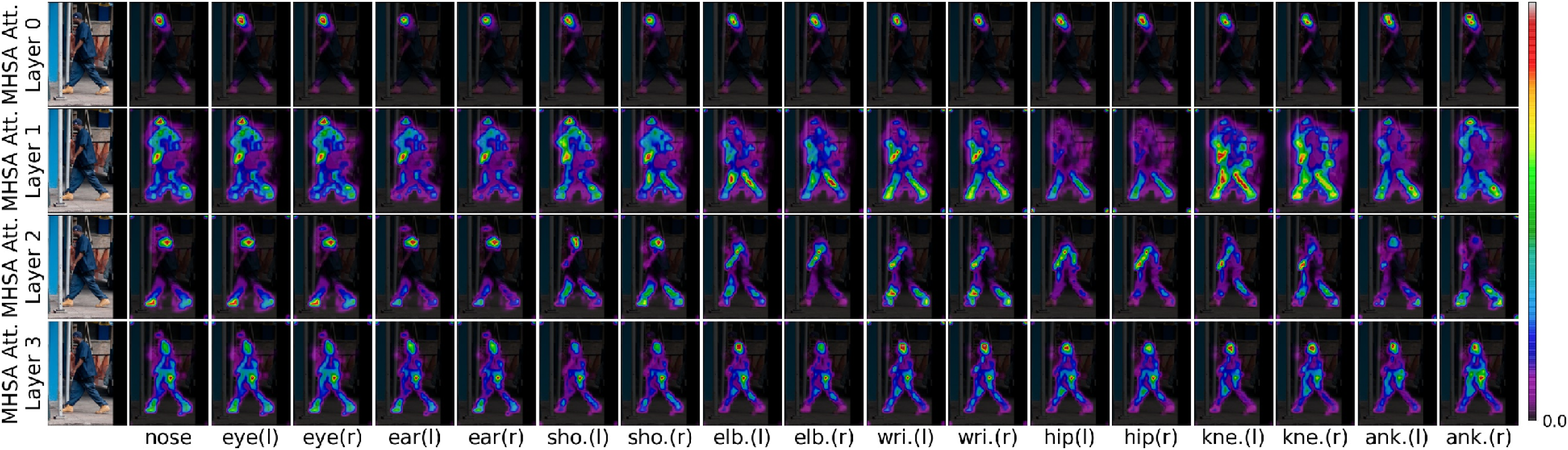} \\
(b)\includegraphics[width=12cm]{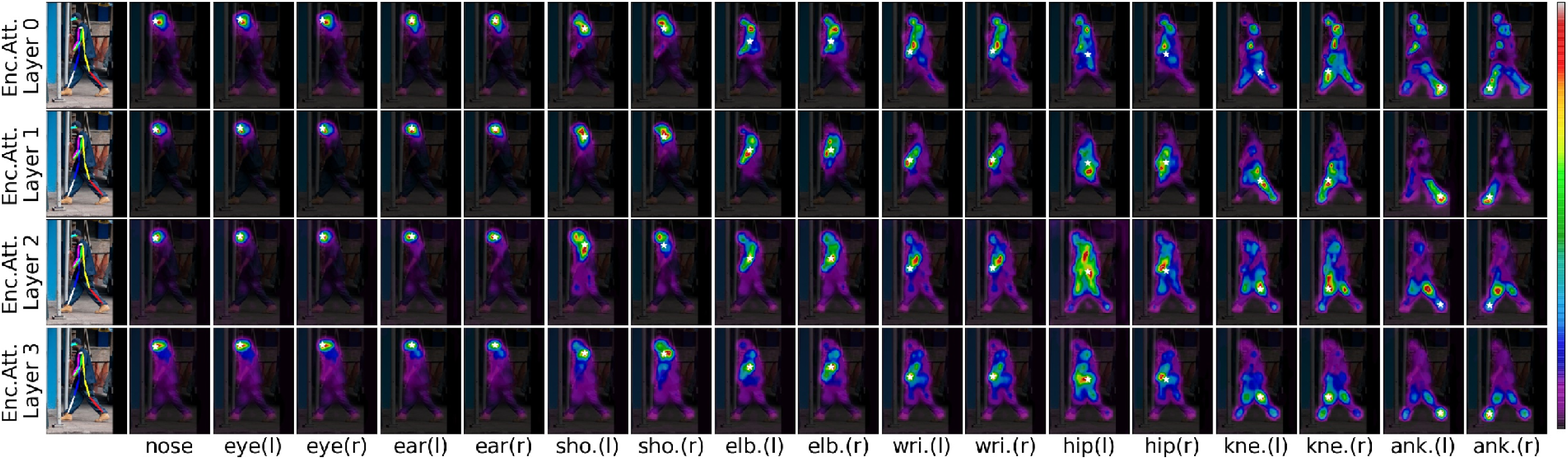} \\
(c)\includegraphics[width=12cm]{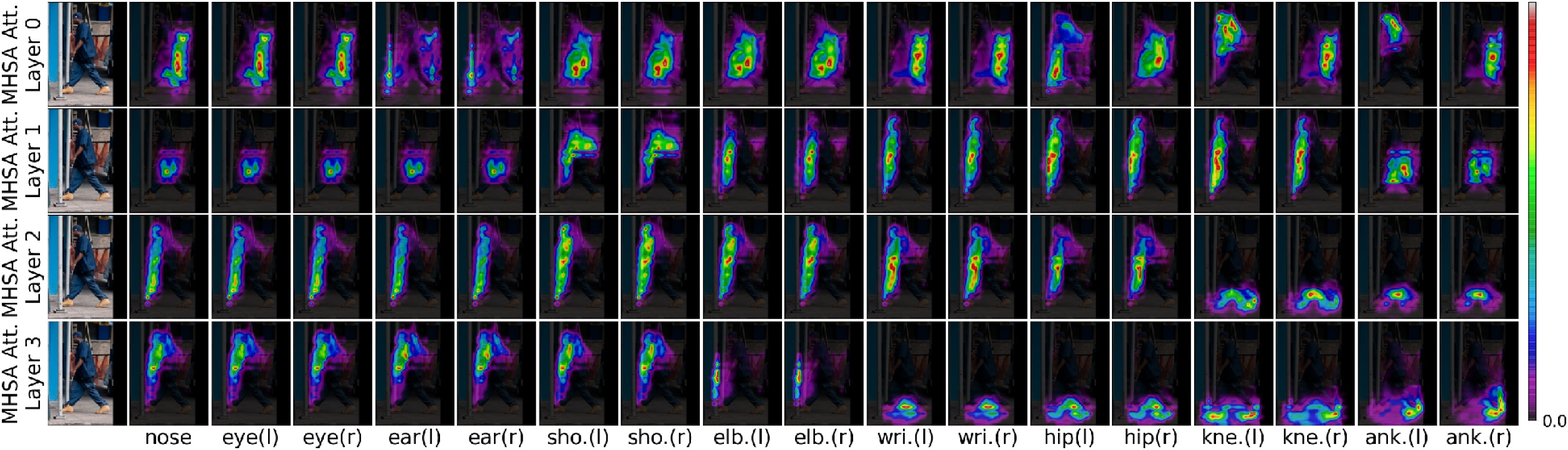}  \\
(d)\includegraphics[width=12cm]{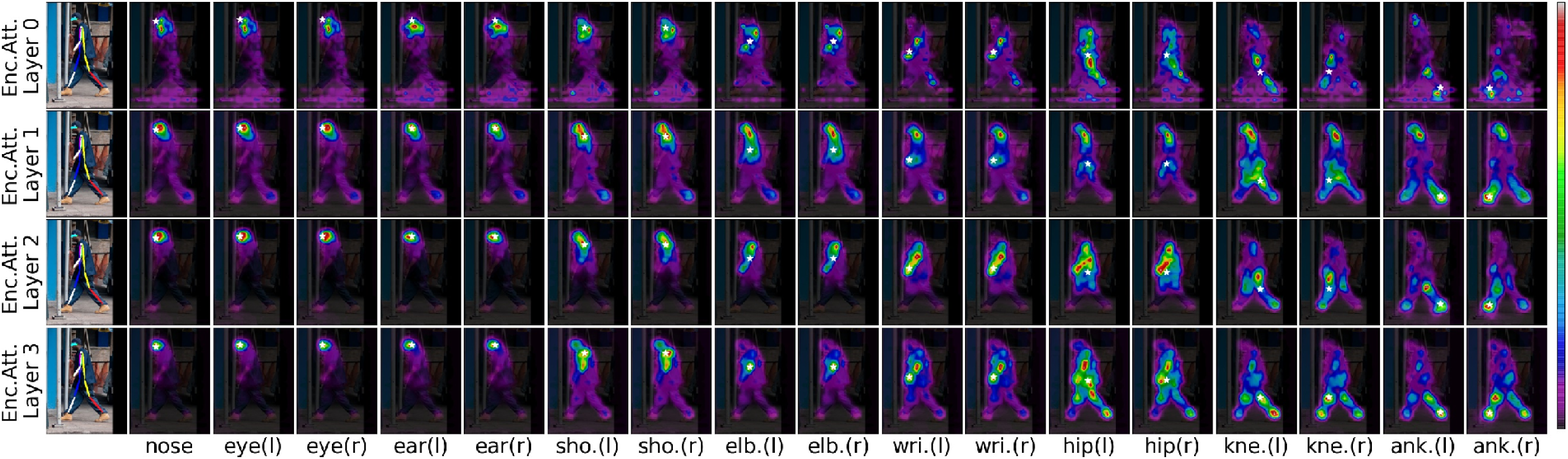} \\ 
\caption{Dependency areas for the predicted keypoints for yet another input image using models: (a \& b) C3A1(4)-Dino; (c \& d) C3A1(4). The dependency area is computed at the MHSA block in (a \& c) and at the Transformer encoder layer in (b \& d).}
\label{fig:da12}%
\end{figure*}

\begin{figure*}[h]
\centering%
(a)\includegraphics[width=12cm]{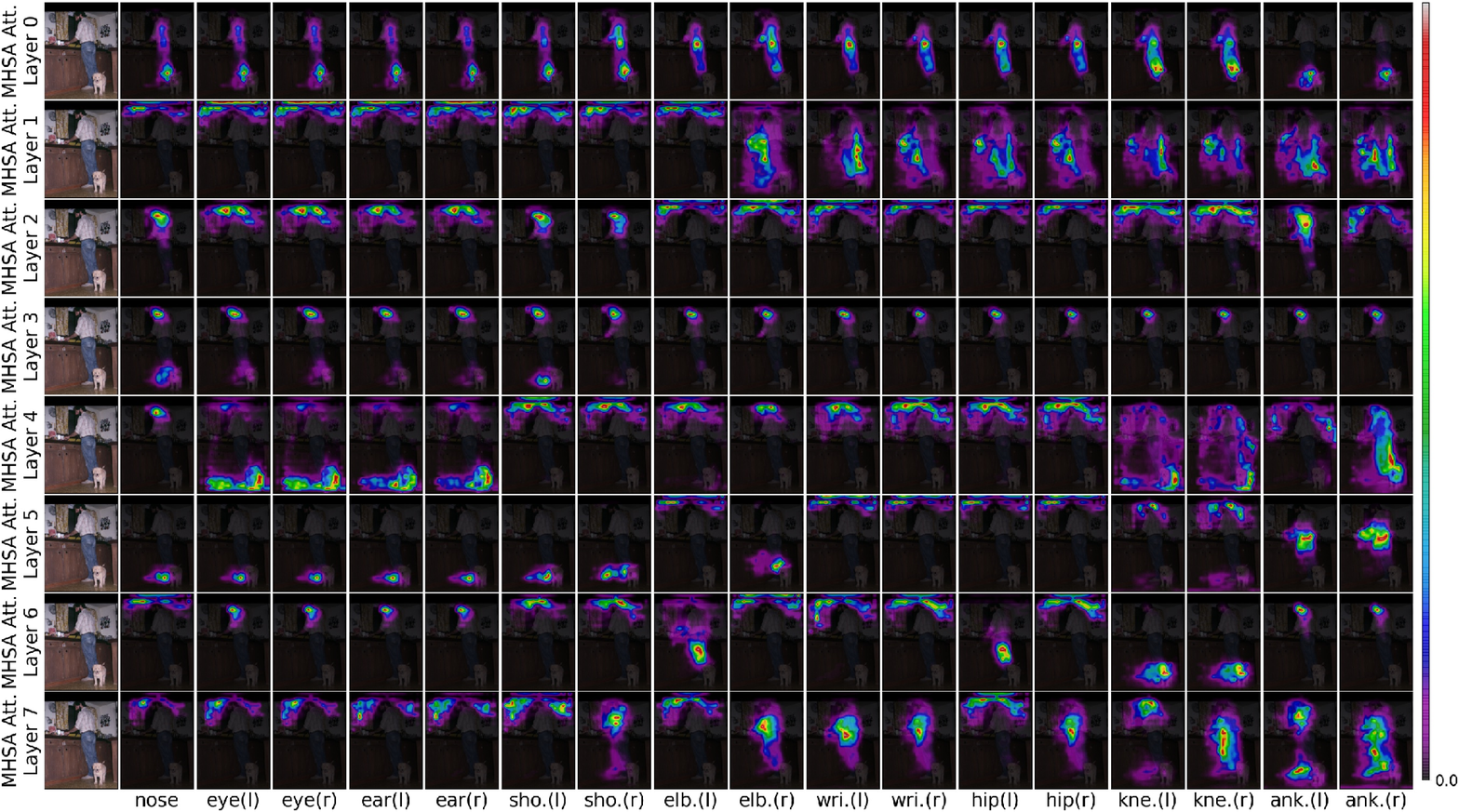} \\
(b)\includegraphics[width=12cm]{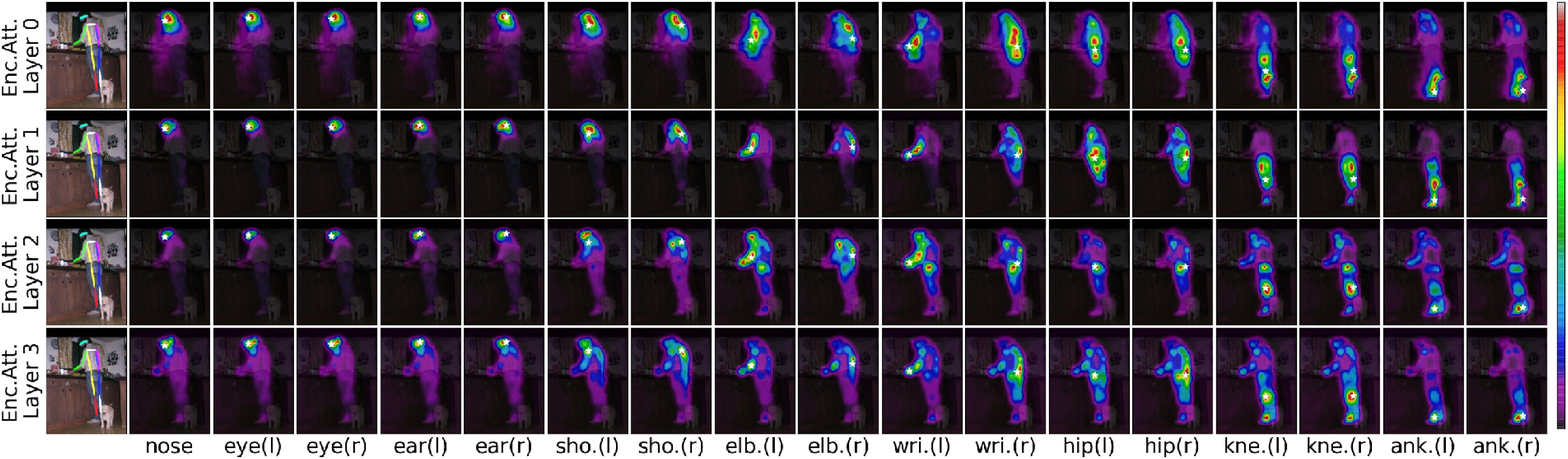} \\
\caption{Dependency areas for the predicted keypoints for the same input image considered in Figure \ref{fig:da11} for C3A1(8)-Dino. The dependency area is computed at the MHSA block in (a) and at the Transformer encoder layer in (b).}
\label{fig:da13}%
\end{figure*}

\begin{figure*}[h]
\centering%
(a)\includegraphics[width=12cm]{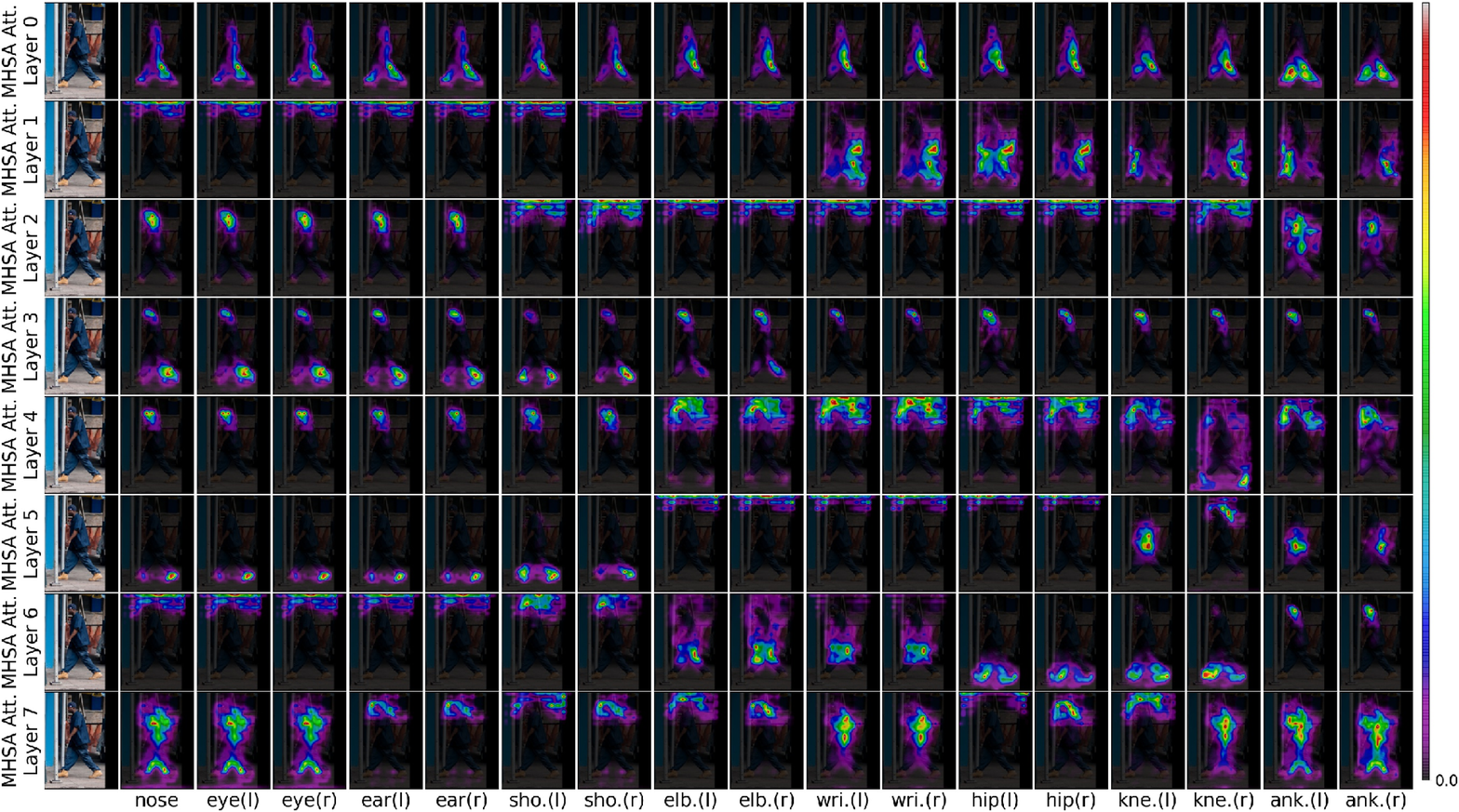} \\
(b)\includegraphics[width=12cm]{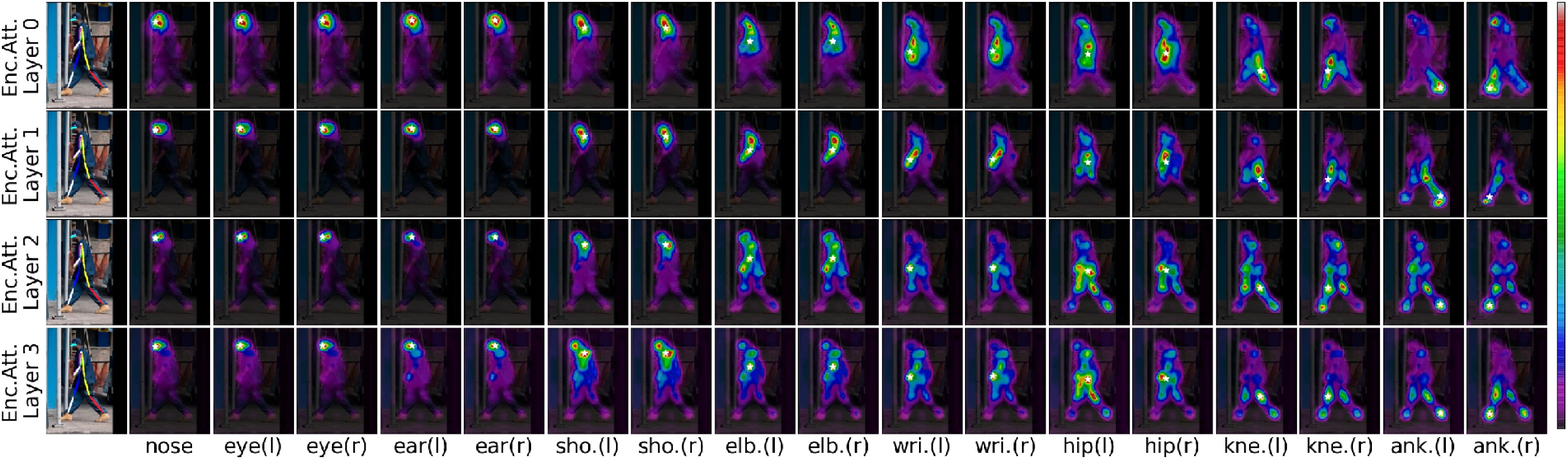} \\
\caption{Dependency areas for the predicted keypoints for the same input image considered in Figure \ref{fig:da12} for C3A1(8)-Dino. The dependency area is computed at the MHSA block in (a) and at the Transformer encoder layer in (b).}
\label{fig:da14}%
\end{figure*}

We consider more input images and compare C3A1(4)-Dino and C3A1(4) to better understand the effect of the self-supervised pre-training and
present the dependency areas in Figures \ref{fig:da11} \& \ref{fig:da12}. 
In Figure \ref{fig:da11}, we consider an image with an occlusion (a dog occluding a human ankle). Here, C3A1(4)-Dino does better than C3A1(4) 
near the occluded region, where the latter is confused (see Figure \ref{fig:da11} b \& d near the ankle regions).
For the same input images, we present the results with C3A1(8)-Dino in Figures
\ref{fig:da13} \& \ref{fig:da14}.

\section*{D. Special scenarios}

\begin{figure*}[h]
\begin{tabular}
[c]{cc}%
\centering%
(a)\includegraphics[width=5cm]{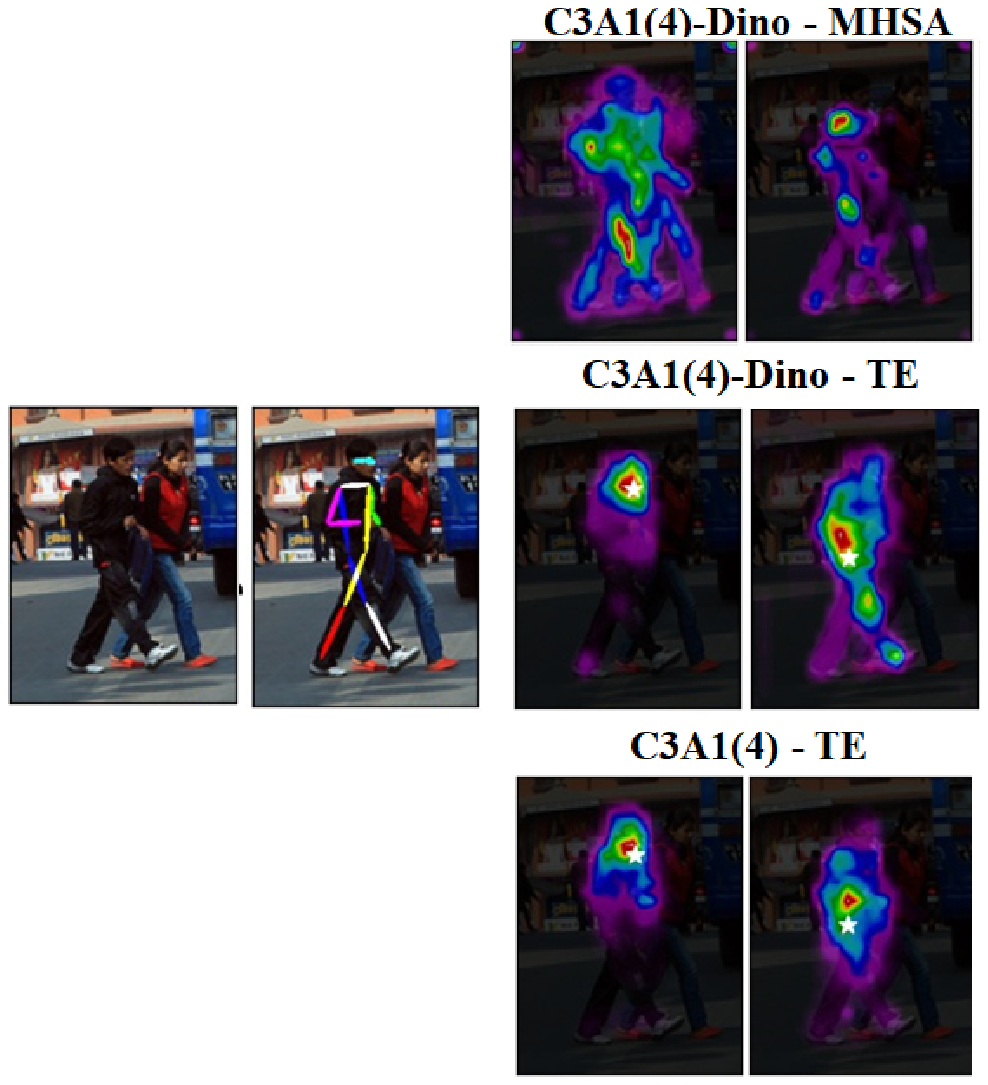} & (b)\includegraphics[width=5cm]{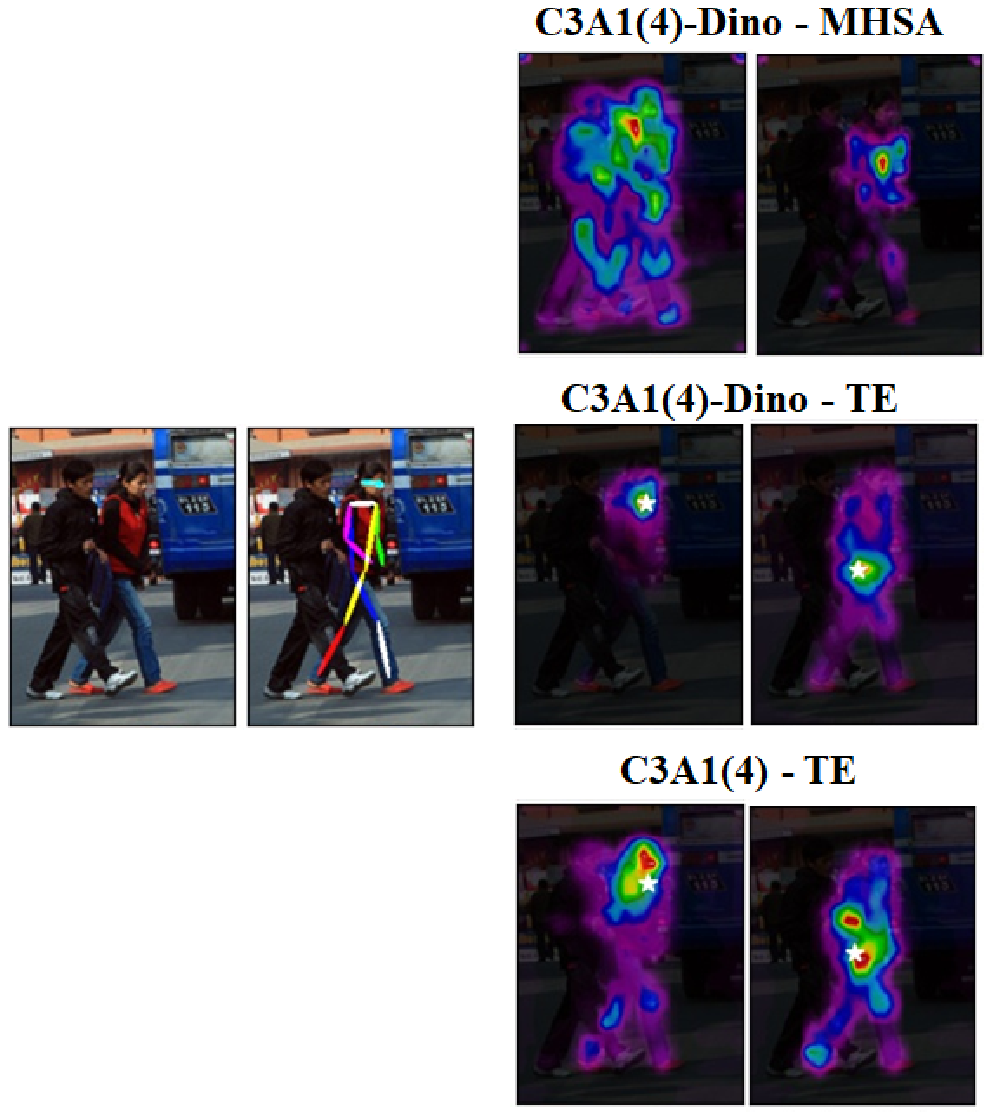} \\
\end{tabular}
\caption{Dependency areas when two humans are close to each other in the input image.}
\label{fig:da22}%
\end{figure*}

{\bf Two humans.} We consider an input image with two humans very close to each other in Figure \ref{fig:da22}. Here, Figure \ref{fig:da22} (a) focuses on the human on the left; and (b) to the 
human on the right. At the MHSA block, we see a large patch that encompasses both humans, suggesting that at the mid-level while the network has figured out the rough region of interest, it is
still unable to distinguish between the two humans and so focuses on both humans. However, at the TE layer the network has figured out the correct human to focus on in each of the input
images. Again, we find the dependency area to more confident for C3A1(4)-Dino (i.e., less pinky zones), and so SSL pre-training makes the network be more confident on where to focus on in the image.   

\end{document}